\documentclass[conference]{IEEEtran}
\IEEEoverridecommandlockouts
\usepackage{url}
\usepackage{hyperref}
\usepackage{cite}
\usepackage{amsmath,amssymb,amsfonts}
\usepackage{algorithmic}
\usepackage{graphicx}
\usepackage{textcomp}
\usepackage{subfig}
\usepackage{xcolor}
\usepackage{cuted}
\def\BibTeX{{\rm B\kern-.05em{\sc i\kern-.025em b}\kern-.08em
    T\kern-.1667em\lower.7ex\hbox{E}\kern-.125emX}}
\begin{document}

\title{PokéAI: A Goal-Generating, Battle-Optimizing Multi-agent System for Pokémon Red
}

\author{Zihao Liu$^{1}$ and Xinhang Sui$^{2}$ and Yueran Song$^{3}$ and Siwen Wang$^{4}$

\thanks{*This work was not supported by any organization.}
\thanks{$^{1}$Zihao Liu is with Qingdao Academy, Qingdao, China. {\tt\small liuzihao\_lzh@foxmail.com}}
\thanks{$^{2}$Xinhang Sui is with Qingdao Academy, Qingdao, China. {\tt\small suixinhang75@gmail.com}}
\thanks{$^{3}$Yueran Song is with Qingdao Academy, Qingdao, China. {\tt\small yueransong0124@gmail.com}}
\thanks{$^{4}$Siwen Wang is with the Department of Mathematics and Computer Science, Qingdao Academy, Qingdao, China. {\tt\small wangsiwen@qdzx.net}}

}

\maketitle

\begin{abstract}
We introduce PokéAI, the first text-based, multi-agent large language model (LLM) framework designed to autonomously play and progress through \textit{Pokémon Red}. Our system consists of three specialized agents—Planning, Execution, and Critique—each with its own memory bank, role, and skill set. The Planning Agent functions as the central brain, generating tasks to progress through the game. These tasks are then delegated to the Execution Agent, which carries them out within the game environment. Upon task completion, the Critique Agent evaluates the outcome to determine whether the objective was successfully achieved. Once verification is complete, control returns to the Planning Agent, forming a closed-loop decision-making system.

As a preliminary step, we developed a battle module within the Execution Agent. Our results show that the battle AI achieves an average win rate of 80.8\% across 50 wild encounters, only 6\% lower than the performance of an experienced human player. Furthermore, we find that a model’s battle performance correlates strongly with its LLM Arena score on language-related tasks, indicating a meaningful link between linguistic ability and strategic reasoning. Finally, our analysis of gameplay logs reveals that each LLM exhibits a unique playstyle, suggesting that individual models develop distinct strategic behaviors.
\end{abstract}


\section{Introduction}
Designing generally capable agents that can reason, plan, and adapt in open-ended environments remains a core challenge in AI \cite{wang2023voyager}. Traditional methods such as reinforcement learning \cite{pleines2025pokemonredreinforcementlearning} or imitation learning \cite{9889671} often rely on low-level actions and hand-crafted rewards, making them brittle and difficult to adapt. Recent progress in large language models has enabled more flexible agents that use natural language as both interface and reasoning substrate. However, many of these systems depend on multi-modal inputs (e.g., vision), which are computationally expensive and difficult to scale—especially in applications requiring many simultaneous agents.

In this work, we introduce PokéAI, the first fully text-based, multi-agent LLM framework designed to autonomously play \textit{Pokémon Red}. Our architecture consists of three specialized agents—Planning, Execution, and Critique—operating in a closed loop. Each agent has a distinct role, memory, and objective, allowing the system to generate goals, carry out game actions, and verify outcomes entirely through text.

We choose \textit{Pokémon Red} as our testbed due to its rich environment, turn-based structure, and demand for long-term planning. While several prior works have explored Pokémon automation, they have predominantly relied on multi-modal pipelines. For instance, ``Gemini plays Pokémon" \cite{Gemini_Plays_Pokemon} and ``Claude plays Pokémon" \cite{ClaudePlaysPokemon} use vision-language models to beat or progress far into the game. Nunu AI \cite{NunuAI} similarly used a multi-modal LLM to reach the third gym in Pokémon Emerald. In contrast, our work is the first to demonstrate a fully text-based, open-source agent framework capable of progressing through the Pokémon game series. Our work is available at: \url{https://github.com/siw028/AI-Pokemon-Trainer}

\section{Method}

\begin{figure}[!ht]
  \centering
  \includegraphics[width=0.48\textwidth]{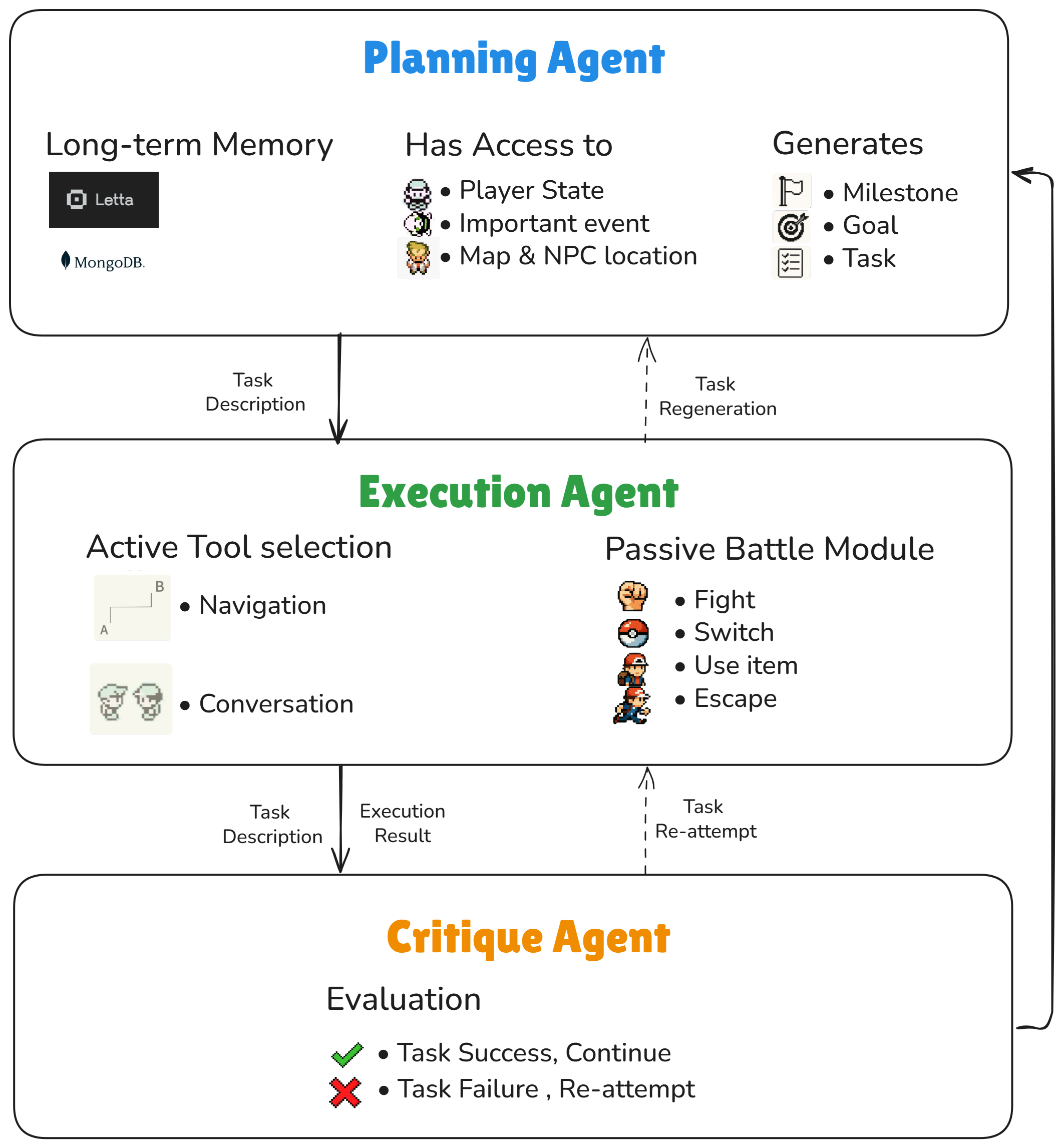}
  \caption{PokéAI Overall Framework}
  \label{fig:framework}
\end{figure}

PokéAI consists of three collaborative agents—the Planning Agent, Execution Agent, and Critique Agent—all powered by a large language model (Fig.~\ref{fig:framework}). Together, they form a closed-loop system for planning, executing, and validating tasks in Pokémon Red.

The Planning Agent is responsible for high-level decision-making. It receives critical game state information, such as the current player location, and retrieves long-term contextual knowledge from a vector-based memory bank. Using this information, it generates a milestone (e.g., beating the next gym), breaks it down into intermediate goals, and finally outputs a sequence of tasks to achieve each goal. This process is iterative: the Planning Agent continuously refines its strategy as new information is added to memory, allowing it to dynamically adapt to the game state.

Once a task is generated, it is passed to the Execution Agent as a string in json format. The Execution Agent first checks whether the task can be completed using its internal toolkit, also stored in a vector memory base. If no suitable tool is found, it sends a “regenerate” request back to the Planning Agent to revise the task. The Execution Agent is equipped with a set of predefined tools—such as a navigation tool that generates a sequence of key presses to move from the current location to a specified destination—which can be invoked through the language model’s function-calling capabilities \cite{wang2025function}. If a battle is encountered during task execution, a passive battle module is triggered automatically. This is detected by monitoring memory address \texttt{0xD057}; when the rightmost bit switches from 0 to 1, it indicates the onset of a battle. Once the battle concludes, the Execution Agent resumes the interrupted task until it is complete.

After finishing a task, the Execution Agent sends an acknowledgment to the Critique Agent, which serves as a task verifier. It receives the original task description, the resulting game state, and an execution summary. For example, if the task was to move from coordinate (1,5) to (6,10), the Critique Agent checks whether the player has reached the destination. If not, it instructs the Execution Agent to retry the task. Once the task is verified as complete, control returns to the Planning Agent, and the cycle continues. 


\section{Preliminary Results}
\begin{figure}[!ht]
  \centering
  \includegraphics[width=0.48\textwidth]{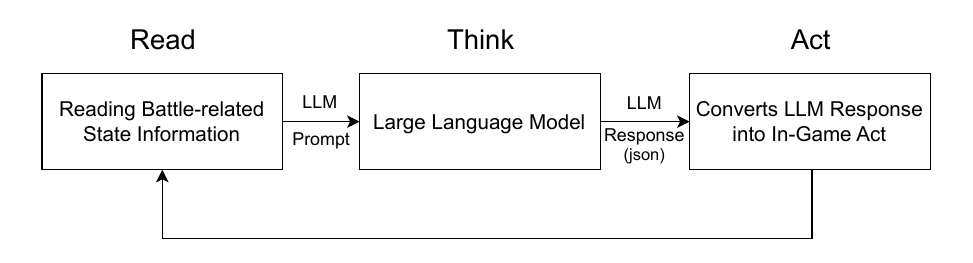}
  \caption{Battle Module flowchart}
  \label{fig:battle flowchart}
\end{figure}

We began assembling PokéAI by first designing the battle module within the Execution Agent, as it is the most self-contained component and functions independently of the other agents. It operates by monitoring a specific memory location in the game to detect the occurrence of a battle. As illustrated in Fig. \ref{fig:battle flowchart}, the battle module comprises four main stages: (1) reading battle-related state information from memory, (2) sending this information to the LLM, (3) receiving the LLM's response, and (4) acting based on the received response. This forms a closed-loop decision-making system that continuously operates until the battle concludes.

To evaluate the effectiveness of our design, we deployed the battle AI in various combat scenarios and tested its performance. We then conducted an ablation study to assess the contribution of each individual component (e.g., Pokémon switching, escaping, using items). Additionally, we evaluated the battle AI with different LLMs as backends, comparing their performance and play style. Finally, we carried out a pilot study to investigate whether the battle module benefits from incorporating long-term memory.


\subsection{Battle Module Performance Evaluation}
We evaluate the battle AI's performance by placing it in Mt.~Moon and allowing it to engage in 50 wild encounter battles. The evaluation uses a game checkpoint featuring a Level 15 Charmander with the moves Scratch, Growl, Ember, and Leer, a Level 15 Pidgey with Gust, Sand Attack, and Quick Attack, and five potions in the bag. Mt.~Moon features four wild Pokémon: Zubat, Geodude, Paras, and Clefairy, with encounter rates of 79\%, 15\%, 5\%, and 1\%, respectively \cite{Altissimo's_Pokémon_Site}. The average level of encountered Pokémon is 8.18.

The experiment is repeated 10 times, and the results are averaged. Our battle agent wins an average of 40.4 out of 50 battles, yielding an 80.8\% win rate. For comparison, we asked an experienced Pokémon enthusiast to play the same scenario. They managed to defeat 43 wild Pokémon before both Charmander and Pidgey fainted, corresponding to an 86\% win rate.

These results suggest that our system achieves performance comparable to that of an experienced human player in simple PvE (player-versus-environment) battles. Note that this evaluation uses DeepSeek-V3 as the backend LLM. As we will show in the following sections, different models produce varying results, although none have yet outperformed an experienced human player.

\subsection{Ablation Study}
We conducted a series of ablation studies to quantify the contribution of each submodule—specifically, Pokémon switching, item usage, and escape—within the battle agent. All evaluations in this study use DeepSeek-V3 as the backend LLM.

As shown in Fig.4, the full-featured battle agent achieved the highest performance, with an average win rate of 80.8\%. The “No Escape” variant followed closely at 79.6\%, indicating that the ability to flee has only a marginal effect on overall success. Interestingly, the error bars for the “No Escape” variant are wider than those of the full-function model, suggesting greater variance in performance. In several trials, the “No Escape” variant even outperformed the full model. This can be interpreted through the lens of The Art of War by Sunzi: “Do not pursue a desperate enemy”\cite{tzu2017art}. In practice, we observed that the full model sometimes overestimated danger and chose to flee unnecessarily. In contrast, disabling escape forced the agent to fight to the end, which occasionally resulted in more wins. However, this strategy also led to greater losses in cases where strategic retreat would have preserved resources—such as encounters with powerful enemies like high-level Geodude.

Next is the “No Switch" variant, which yielded a win rate of 58.8\%. Note that in this condition, only strategic switching is disabled; the AI may still switch Pokémon if one faints. The significant performance drop highlights the model's ability to engage in forward planning and select the optimal Pokémon for each situation.

Finally, the “No Item” variant performed the worst, with an average win rate of only 32.6\%. This result aligns with expectations, as removing the five healing potions severely handicaps the agent's ability to sustain its team during extended combat.

\begin{figure}[!ht]
  \centering  \includegraphics[width=0.48\textwidth]{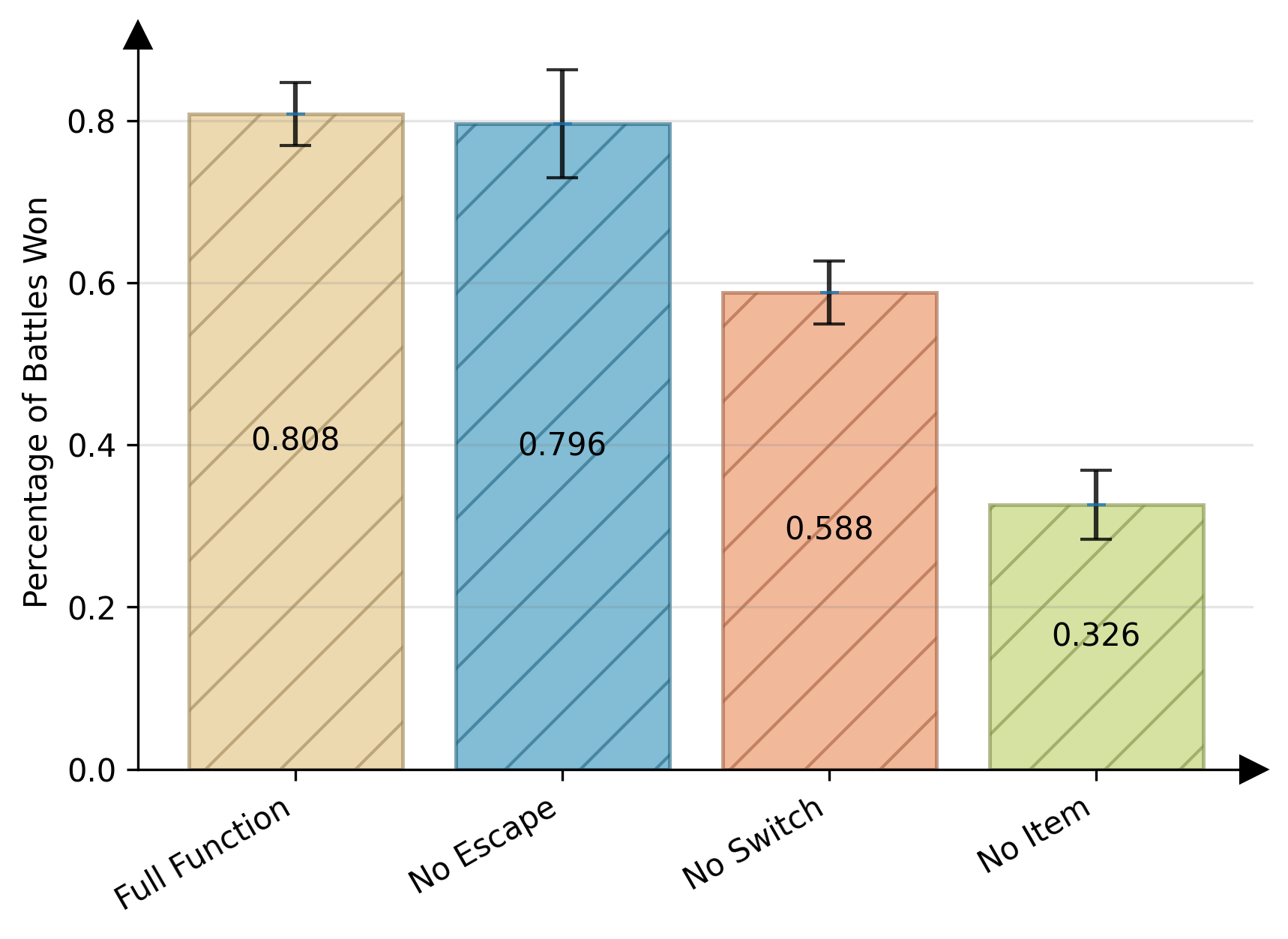}
  \caption{Ablation study on action availability and win rate. Error bars show ±1 standard error of the mean}
  \label{fig:ablation res}
\end{figure}

\subsection{LLM Performance Comparison}
In addition to the ablation study, we also compare the battle agent’s performance across different LLMs in the backend. As a baseline, we use a randomized model that selects from available button inputs with equal probability. To avoid dead ends, we only sample from dynamically valid in-game actions.

\begin{figure}[!ht]
  \centering  \includegraphics[width=0.48\textwidth]{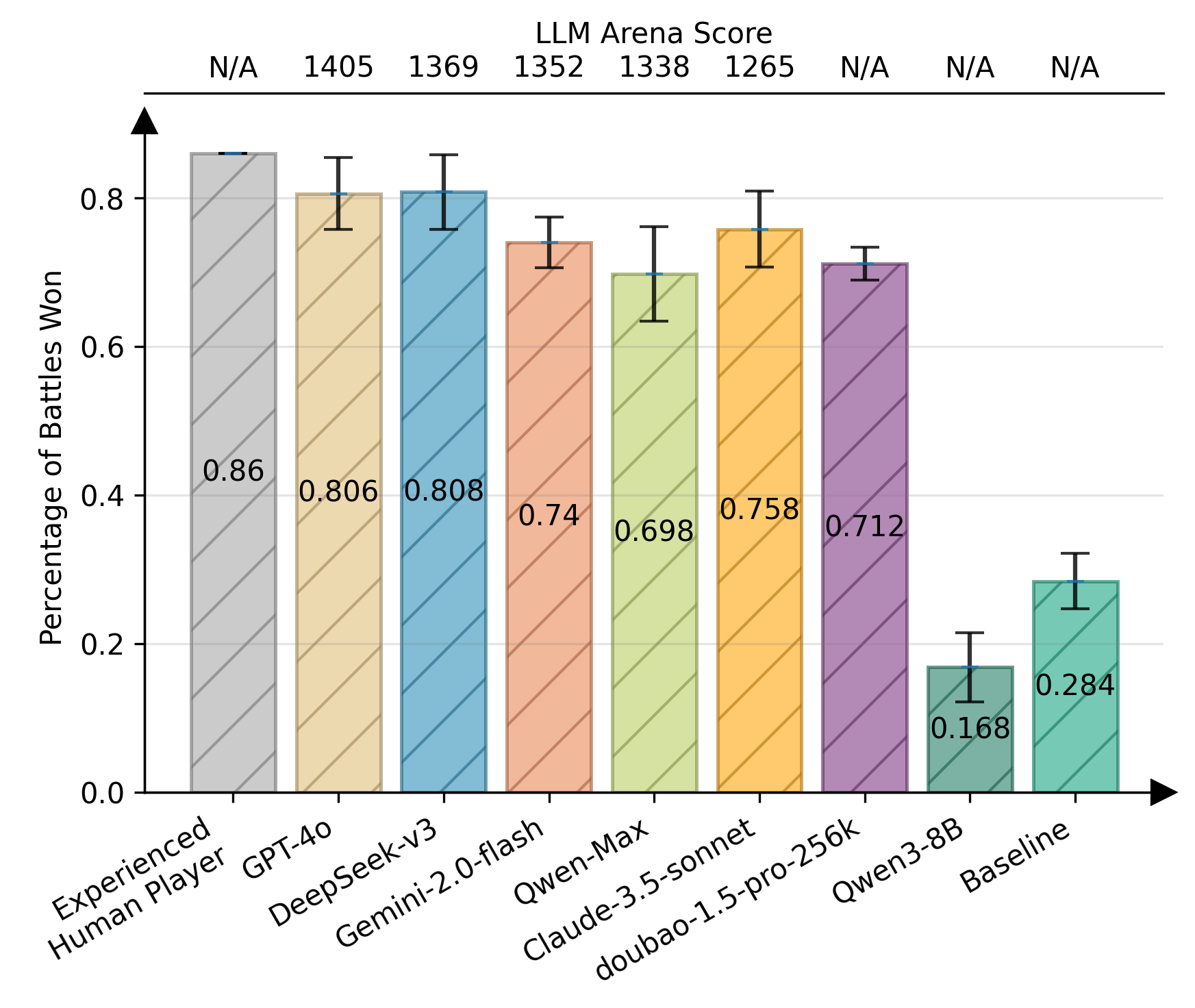}
  \caption{Performance comparison across LLMs and experienced human player. Error bars show ±1 standard error of the mean}
  \label{fig:comparison across LLMs}
\end{figure}

As shown in Fig.~\ref{fig:comparison across LLMs}, we observe that model performance is generally proportional to the LLM Arena Score for language-related tasks, with the exception of Claude 3.5 Sonnet, which outperforms some models with higher scores. Due to time and budget constraints, we did not fully evaluate some reasoning-oriented models like DeepSeek-R1. Preliminary tests suggest that reasoning ability does not provide a significant edge in this simple battle scenario—on the contrary, it may even lead to worse decisions, illustrating the downside of overthinking \cite{cuadron2025danger}.


\subsection{LLM Playstyle Analysis}
Prior research suggests that LLMs often exhibit distinct "personalities" \cite{serapio2023personality}, and efforts in AI safety have focused on aligning LLM behavior with desirable human traits \cite{li2024big5}. Motivated by this, we analyze whether different LLMs exhibit different battle playstyles. Figure~\ref{fig:LLM playstyle analysis} shows the distribution of actions taken during battles, helping explain performance discrepancies observed in the previous section.

Qwen-3B is the only model that performs worse than the randomized baseline. It often selects ineffective moves (e.g., repeatedly using non-damaging moves) and tries to use moves with no remaining Power Points (PP), causing the game to get stuck and runs to terminate early.

Doubao-1.5-pro-256K performs modestly but demonstrates highly consistent behavior with minimal variance. Its action distribution closely resembles that of an experienced human player, although it switches Pokémon less frequently. It also occasionally wastes potions on Pokémon that have no usable moves left, indicating a lack of contextual awareness.

Claude 3.5 sonnet achieves a solid win rate of 75.8\% and displays a distinctive playstyle not observed in other LLMs. Once battle starts, Claude tends to make a single switch in the very first round to choose the most optimal Pokémon based on the current opponent and commits to that choice for the rest of the match. 

GPT-4o and DeepSeek-v3 both achieve high win rates (80.6\% and 80.8\% respectively) but with markedly different styles. GPT-4o avoids non-damaging moves and rarely escapes from battles, favoring an aggressive strategy. It even achieved a perfect 50/50 win rate in some runs. However, its reluctance to switch Pokémon or escape often leads to quicker HP depletion. DeepSeek-v3, on the other hand, makes frequent switches and escapes, demonstrating a more strategic and resource-preserving approach. It appears to consider type advantages and risk more carefully.

A particularly curious behavior emerged with GPT-4o. Despite being instructed explicitly to only use Item 4 (HP-restoring potion), it occasionally attempts to use Item 1 (Pokéball) when both Pokémon are low on HP—an apparent attempt to avoid further damage. This suggests GPT-4o may be engaging in “creative rule-bending”, interpreting the presence of other items in the prompt as a license to improvise \cite{murthy2024evaluating} \cite{meinke2024frontier}.
\begin{strip}
  \centering
  \includegraphics[width=\textwidth]{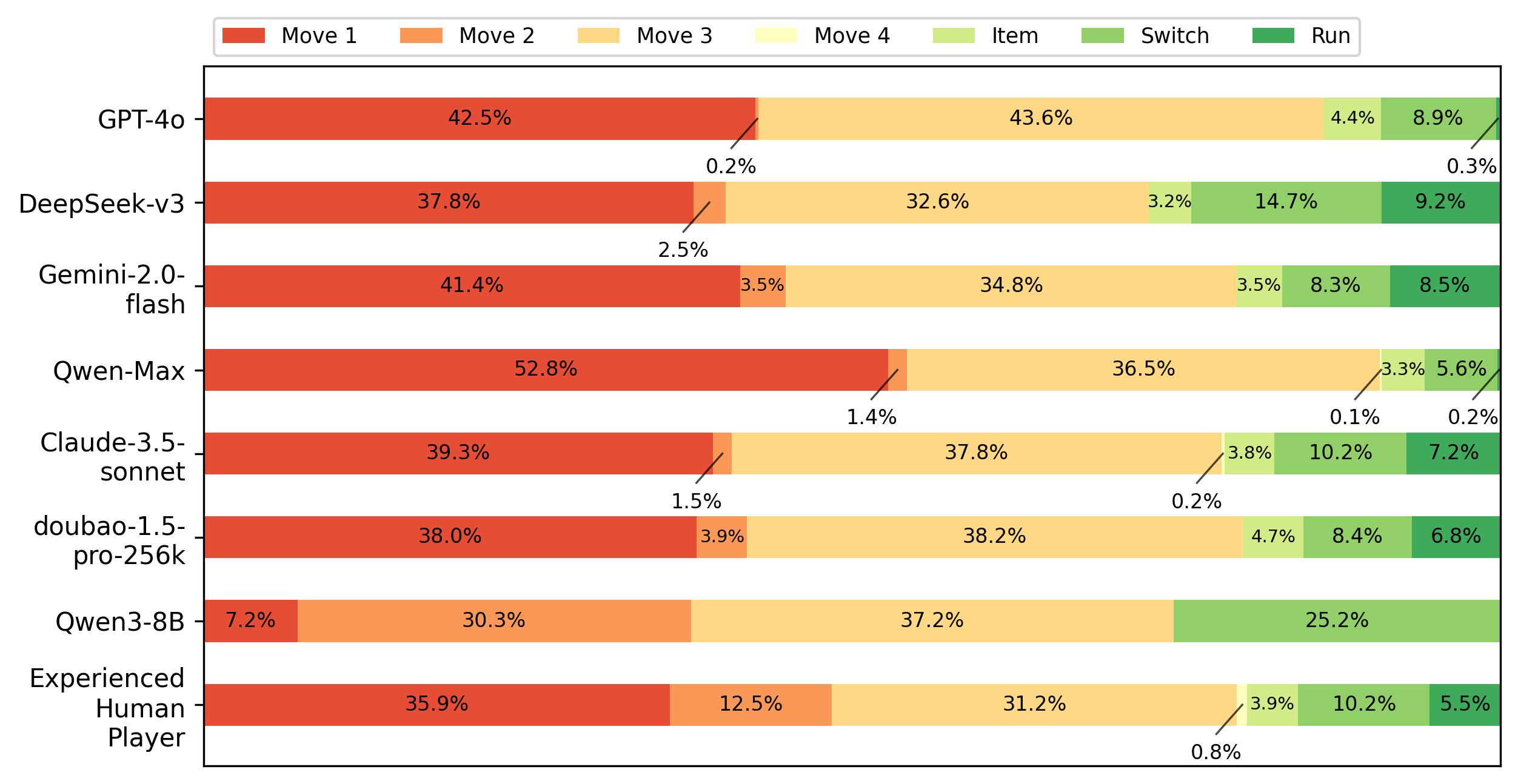}
  \captionof{figure}{Action Distribution by LLM Models and Experience Human Player During Battles}
  \label{fig:LLM playstyle analysis}
\end{strip}





\subsection{Long-term memory with Letta Agent Framework}
Our current battle module relies solely on short-term memory, limited to the operation history of the last three rounds. However, incorporating long-term memory—such as records of past successes and failures—may further enhance decision-making. To explore this possibility, we conducted a pilot study using the Letta Agent framework \cite{packer2023memgpt}.

Letta is an agent development framework that enables direct access to and manipulation of an agent’s memory. It uses ``heartbeat events" as a mechanism for recalling, chaining, and modifying internal memory traces, allowing agents to reflect on prior experiences during task execution.

To evaluate the impact of long-term memory on battle decisions, we injected a memory snippet into Letta’s memory bank: \textit{“Level 5 Squirtle was defeated by a Level 8 Pikachu in Viridian Forest.”} We then simulated a new wild encounter, this time pitting a Level 6 Squirtle against a Level 9 Pikachu. The goal was to observe whether the agent would recall the prior loss and opt to flee from the battle.

\begin{figure}[!ht]
  \centering
  \includegraphics[width=0.48\textwidth]{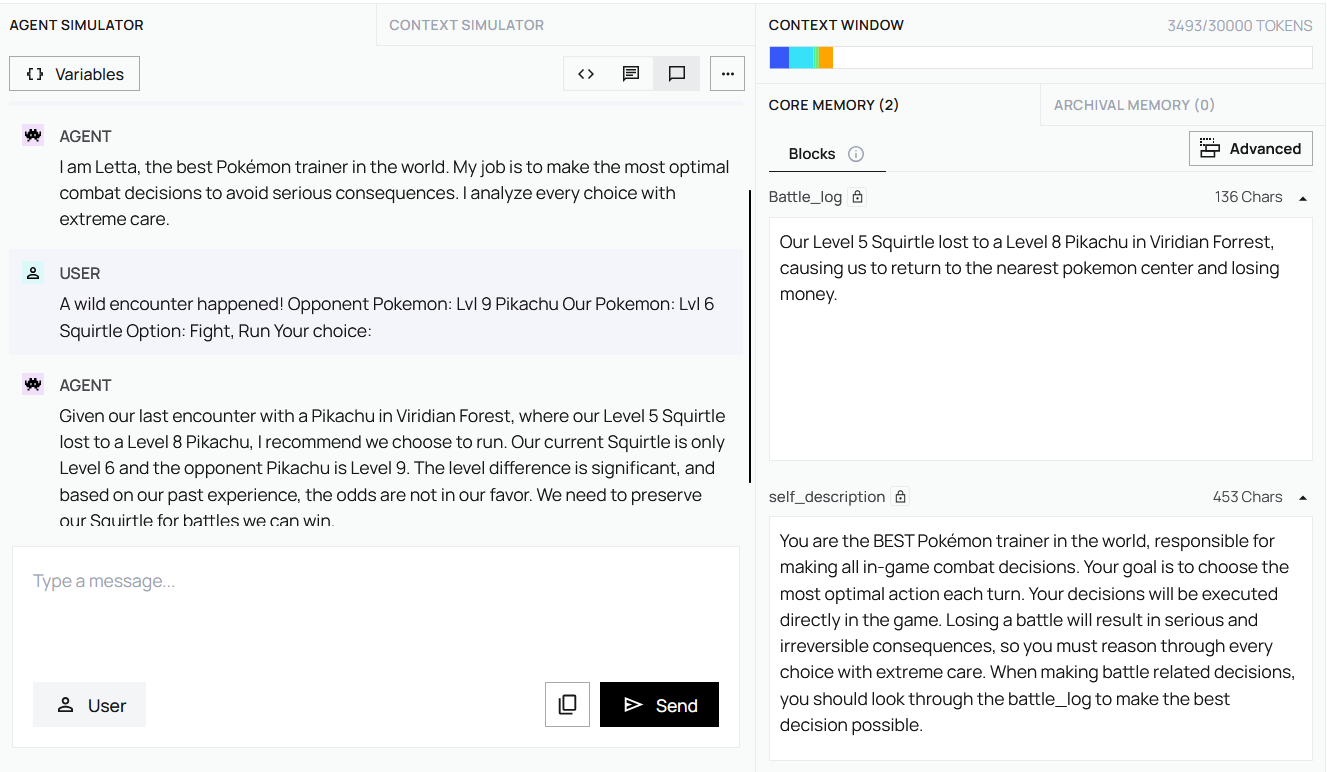}
  \caption{Long-term memory with Letta Agent, past battle logs improve future decision making}
  \label{fig:letta}
\end{figure}

The results in Fig. \ref{fig:letta} show that the Letta agent successfully retrieved the injected memory, reasoned about the similarity of the current situation, and chose to run, correctly identifying it as the safer option. This preliminary experiment suggests that integrating long-term memory into agent design can improve strategic decision-making and overall battle performance.

\section{Future Work}
Our current work focuses on the battle module within the Execution Agent. As immediate future work, we plan to implement and integrate the rest of the Execution Agent, the Planning Agent, and the Critique Agent to complete the closed-loop decision-making system.

Once the agent is fully functional,  in future iterations, we aim to explore more open-ended, simulation-based approaches. For example, we plan to initialize agents with different personality traits or gameplay preferences to examine how these factors influence decision-making and playstyle. This could help us understand whether diverse agent profiles lead to emergent behavior, providing insights into modeling more human-like, adaptive NPCs in game environments.




\bibliographystyle{unsrt}  
\bibliography{ref}  

\end{document}